\pdfoutput=1

\documentclass[11pt]{article}

\usepackage{coling}

\usepackage{times}
\usepackage{latexsym}

\usepackage[T1]{fontenc}

\usepackage[utf8]{inputenc}

\usepackage{microtype}

\usepackage{inconsolata}

\usepackage{amsmath}
\usepackage{amssymb}
\usepackage{adjustbox}
\usepackage{booktabs}
\usepackage{tablefootnote}
\usepackage{pifont}
\usepackage{svg}
\usepackage{makecell}
\usepackage{enumitem}
\usepackage{fvextra}
\usepackage{fancyvrb}
\usepackage{spverbatim}

\title{Language Models Learn Metadata: Political Stance Detection Case Study}

\author{Stanley Cao \\
  Department of Computer Science \\
  Stanford University \\
  \texttt{stanley.l.cao@stanford.edu}
  \\\And
  Felix Drinkall \\
  Oxford-Man Institute of Quantitative Finance \\
  University of Oxford \\
  \texttt{felix.drinkall@eng.ox.ac.uk} \\
  }

\begin{document}
\maketitle
\begin{abstract}
Stance detection is a crucial NLP task with numerous applications in social science, from analyzing online discussions to assessing political campaigns. This paper investigates the optimal way to incorporate metadata into a political stance detection task. We demonstrate that previous methods combining metadata with language-based data for political stance detection have not fully utilized the metadata information; our simple baseline, using only party membership information, surpasses the current state-of-the-art. We then show that prepending metadata (e.g., party and policy) to political speeches performs best, outperforming all baselines, indicating that complex metadata inclusion systems may not learn the task optimally.
\end{abstract}

\section{Introduction}
Parliamentary debates contain significant information about the political climate, directly influencing national policy decisions. Understanding the stances within parliamentary debates between opposing political parties provides insights into national policies and support networks within the government. Additionally, examining the stance differences within the same party offers an interesting perspective on internal dynamics and ideological diversity. Data about political debates aids citizens in learning more about political representatives \citep{chaffee1980, mckinney2003viewer}. Determining the stances of individual
politicians on motions and policies remains a non-trivial, yet important task, as belonging to the same party does not guarantee shared political views.

Political stance detection involves identifying a speaker's position or viewpoint on a specific motion made by another speaker. This task remains a challenging task for a number of reasons; specific contextual information, esoteric procedural jargon, compound opinions, comparative structures, and non-literal word usage all contribute to the difficulty of this task \citep{edelman1985}. Moreover, political stance detection is a challenging task for humans \citep{thomas-etal-2006-get, Salah2014MachineLA}, which has motivated research in creating automated models for predicting legislator behavior \citep{Poole1985ASM, gerrish_blei_2012, nguyen-etal-2015-tea} and political stances \citep{bhatia-p-2018-topic, bhavan-etal-2019-investigating}.

In the current research literature, several proposed models use textual information and metadata in various ways, ranging from rule-based approaches to sophisticated graph embedding systems such as graph attention networks and mining knowledge graphs from textual data \citep{Onyimadu2013TowardsSA, Chen2017OpinionawareKG, sawhney-etal-2020-gpols, Bhavan2020AnalysisOP, davoodi-etal-2022-modeling}. Notably, \citet{TEC_Sawhney} propose a time-evolving graph-based model that incorporates the temporal relationships between different politicians for the political stance detection task. Work specifically focusing on the incorporation of metadata in political contexts has been done \citep{linder_2020, yano-etal-2012-textual, kornilova-etal-2018-party}, showing that manual metadata feature-engineering can augment legislative texts for better model performance.

However, much of the existing research has either overlooked metadata or failed to utilize it in a scalable and effective manner. This paper shows the sensitivity of different models to various components of metadata, culminating in three main contributions:
\begin{enumerate}[itemsep=-1pt, topsep=1pt]
    \item We propose a simple Bayesian model that uses metadata only, yet 
outperforms all models introduced in prior work on the \textit{ParlVote+} dataset, demonstrating that metadata is being fundamentally underutilized.
    \item Previous approaches have overcomplicated the task, as we prove that a simple prepending mechanism can better 
incorporate metadata for this task, suggesting 
that language models can successfully incorporate metadata without having to explicitly construct models or manufacture features based on metadata.
    \item Small fine-tuned encoder-based language models outperform larger generative language models in a zero-shot setting \citep{brown2020, chatgpt4o_2024}.
\end{enumerate}



\section{Related Work}
Stance detection has emerged as a significant focus in NLP research \citep{Kk2020StanceD, Hardalov2021ASO}, notably applied to political debates with the creation of the \textit{ConVote} dataset by
\citet{thomas-etal-2006-get}. This dataset comprises labelled US congressional floor debates, annotated according to the voting behaviour of the speakers.
Similarly, the \textit{HanDeSet} dataset \citep{abercrombie-batista-navarrow-2018-hansard} includes transcripts from the UK Parliament, featuring two types of polarity labels --- one set of labels is based on the speakers' votes, while the other is manually assigned by a human who reviews the speech and the motion to determine the label.
\citet{abercrombie-batista-navarro-2020-parlvote, abercrombie-batista-navarro-2022-policy} improve upon this, creating
the \textit{ParlVote} dataset comprising of 34,010 examples of parliamentary debates between 1997-2019, and a 
revised dataset
\textit{ParlVote+} which includes policy preference labels for each example and
removes some incorrect examples from the original dataset. These datasets are particularly useful for studying voting patterns, which are influenced by sophisticated social networks \citep{wojcik}.
As new political speech datasets are created, analyzing and mining opinions from
political discourse has become more popular, \citep{cabrio-villata-2018-five},
especially as large language models (LLMs) become more mainstream \citep{devlin2019bert, Yang2019XLNetGA, Touvron2023LLaMAOA, Enis2024claude}. Recent political sentiment analysis research now includes multiple languages, nations, and communication platforms, reflecting a broad range of contexts where political discourse is important \citep{Binnewitt2024RecognisingOT, Laabar2024MultiDimensionalIA, Miok2022MultiaspectMA, A2023AnalyzingTV}.

In this paper, we focus on the \textit{ParlVote+} dataset. Other datasets curated for political stance detection such as \textit{P-Stance} \citep{p-stance} and \textit{SemEval-2016} \citep{mohammad-etal-2016-semeval} are not as lexically sophisticated as \textit{ParlVote+}, which includes complex political vernacular distinctive to UK Parliamentary speeches.  To the best of our knowledge, the state-of-the-art model for political stance detection on the \textit{ParlVote+} dataset is GPolS
\citep{sawhney-etal-2020-gpols}, which achieves an accuracy of 76\% on the full corpus with a random train/test split.


\section{Methodology}

The problem formulation for analyzing parliamentary debates is exactly that
stated by \citet{sawhney-etal-2020-gpols}. For each motion $m$, we have the
transcript $t$ of the speech made by MP speaker $s$, who is affiliated with political party $p$, in response to motion $m$. The
task is to determine whether $s$ voted in support or in opposition of $m$. 

\subsection{Data}
The \textit{ParlVote+} dataset consists of 33,311
labelled examples of English language parliamentary debates. Each example
contains the speech made by a particular speaker, the motion to which the
the speaker is responding to, and the vote of the speaker (in support or in opposition
to the motion). Additional metadata is included as well, such as party
affiliation of the speaker and motioner. 

Prior work mainly uses two different data splits. In papers such as \citet{abercrombie-batista-navarro-2020-parlvote} and \citet{sawhney-etal-2020-gpols}, the authors use an 80/10/10 train-validation-test split. We use the full dataset, including all parliamentary motions and speeches in our data split. However, \citet{TEC_Sawhney} split the data temporally; this prevents speeches in response to the same motion from being split across train and test sets, which avoids partial voting and policy information from leaking between the data splits. To create the temporal split, we create the train set by including all motions and speeches until 24/11/2015, and randomly splitting the remaining data into the validation and test sets. We chose this particular cutoff date to closely approximate an 80/10/10 split to maintain consistency between the two different data splits. We then evaluate our models on both data splits. 

\subsection{Bayesian Models}
\label{sec:baysian_model}
We generate our initial probability forecast using a Naive-Bayes method that depends
only on the parties of the speaker and motioner. 
We calculate the probability that a speaker from party $s_j$ votes in favor of a motion (i.e., $V = 1)$ from party $m_i$:
\begin{equation}
\label{eqn:party}
\Pr(V = 1 \mid M = m_i, S = s_j).
\end{equation}
In the \textit{ParlVote}+ dataset, each motion and speech pair is also labelled with a policy number, which identifies which policy the original motion discusses. Because certain political parties have different stances on different policies, we create an augmented version of our previous Bayesian model by incorporating policy metadata. 
To incorporate policy information, we compute for a policy $p_k$:
\begin{multline}
    \label{naive_bayes_with_policy}
    \Pr(V = 1 \mid M = m_i, S = s_j, P = p_k).
\end{multline}
Incorporation of policies is based on a t-test to determine whether the policy statistic significantly differs from the global party statistic. See \ref{appdx:ttest}.


\subsection{Finetuning MPNet}
We include textual information by using MPNet \citep{reimers-2019-sentence-bert, Song2020MPNetMA}, a sentence-transformer that improves upon BERT by combining the advantages of masked language modeling and permuted language modeling to train better embeddings. Instead of using only the CLS token for classification, we take the average of the entire final layer to include more information into the sentence embedding. We then pass this embedding to the classification layer and apply the negative log-likelihood loss criterion.



\subsection{Hybrid Models}
To combine metadata information with the textual data from the motions and speeches, we compare two forms of hybrid models.

\subsubsection{Combining Bayesian Probabilities with Sentence Transformers}
This formulation combines the Bayesian probabilities derived from party and policy metadata with the textual information encoded by MPNet. We directly concatenate the probabilities to the sentence embedding, passing this augmented vector through the classification layer.

\subsubsection{Prepending Metadata to Textual Data}
This formulation prepends the party and the policy to the corresponding motion / speech. before the text is passed through MPNet. Since the party and policy metadata are specialized tokens, we allow the language model to interpret the meaning of the party and policy via finetuning. By learning the word embeddings for parties and policies, MPNet uses its attention mechanism to make informed decisions about the stance of the speaker. 

\subsection{GPT-4o}
We benchmark GPT-4o on this task in both a 0-shot and 6-shot setting. Specifically, we use ChatGPT-4o (2024-05-13) for all experiments in this paper. In \ref{chatgpt_prompts}, we show the prompts used for prompting GPT-4o, including the few-shot setup, and the metadata incorporation. For the 6-shot setting, we give the model an equal number of examples that are labeled $V = 0$ and $V = 1$. We also ensure to provide GPT-4o with challenging examples, such as intra-party disagreement, inter-party agreement, and speeches from minority parties. 


\section{Results}
\begin{table}
\footnotesize
\centering
\begin{tabular}{lccccc}
\toprule
\multicolumn{5}{l}{\textbf{Baselines}} \\
Model & Text & \makecell{Part.} & \makecell{Pol.} & \makecell{Rand. \\ Acc.} & \makecell{Temp. \\ Acc.}\\
\midrule
Random & \ding{55} & \ding{55} & \ding{55} & 0.51 & N/A\\
BERT + MLP & \ding{51} & \ding{55} & \ding{55} & 0.65 & N/A \\
Deepwalk & \ding{51} & \ding{51} & \ding{55} & 0.72 & N/A\\
GPolS & \ding{51} & \ding{51} & \ding{55} & 0.76 & 0.74\\
TEC & \ding{51} & \ding{51} & \ding{55} & N/A & \textbf{0.75} \\
Bayesian & \ding{55} & \ding{51} & \ding{55} & 0.80 & 0.73 \\
Bayesian & \ding{55} & \ding{51} & \ding{51} & \textbf{0.81} & 0.74 \\
\bottomrule
\toprule
\multicolumn{5}{l}{\textbf{Bayesian Hybrid Models}} \\
Model & Text & \makecell{Part.} & \makecell{Pol.\footnotemark} & \makecell{Rand. \\ Acc.} & \makecell{Temp. \\ Acc.} \\
\midrule
MPNet + MLP & \ding{51} & \ding{51} & \ding{55} & 0.83 & \textbf{0.77} \\
MPNet + MLP & \ding{51} & \ding{51} & \ding{51} & \textbf{0.87} & \textbf{0.77} \\
\bottomrule
\toprule
\multicolumn{5}{l}{\textbf{Pure Language Models}} \\
Model & Text & \makecell{Part.} & \makecell{Pol.} & \makecell{Rand. \\ Acc.} & \makecell{Temp. \\ Acc.}\\
\midrule
MPNet + MLP & \ding{55} & \ding{51} & \ding{55} & 0.80 & N/A \\
MPNet + MLP & \ding{55} & \ding{51} & \ding{51} & 0.83 & N/A \\
MPNet + MLP & \ding{51} & \ding{55} & \ding{55} & 0.75 & 0.64 \\
MPNet + MLP & \ding{51} & \ding{51} & \ding{55} & 0.86 & 0.76 \\
MPNet + MLP & \ding{51} & \ding{51} & \ding{51} & \textbf{0.88} & \textbf{0.77} \\
\bottomrule
\toprule
\multicolumn{5}{l}{\textbf{GPT-4o}} \\
Model & Text & \makecell{Part.} & \makecell{Pol.} & \makecell{Rand. \\ Acc.} & \makecell{Temp. \\ Acc.}\\
\midrule
0-shot Prompt & \ding{51} & \ding{55} & \ding{55} & 0.66 & 0.66 \\
0-shot Prompt & \ding{51} & \ding{51} & \ding{55} & 0.71 & 0.71 \\
0-shot Prompt & \ding{51} & \ding{51} & \ding{51} & 0.71 & 0.71 \\
6-shot Prompt & \ding{51} & \ding{55} & \ding{55} & 0.64 & 0.64 \\
6-shot Prompt & \ding{51} & \ding{51} & \ding{55} & \textbf{0.72} & \textbf{0.74} \\
6-shot Prompt & \ding{51} & \ding{51} & \ding{51} & \textbf{0.72} & \textbf{0.74} \\
\bottomrule
\end{tabular}
\caption{Overall performance measurements on the random data split and the temporal data split. 
}
\label{overall-stats}
\end{table}

Table \ref{overall-stats} contains the overall performance of our models in comparison to previous state-of-the-art models on the Parlvote dataset.
We observe that models that incorporate meta-data typically outperform their vanilla counterparts. The incorporation of policy data in addition to the party data further increases accuracy for MPNet, suggesting that both policy and party data are useful for prediction after finetuning. We also note that party information is quite helpful for GPT-4o for inference in both the 0-shot and 6-shot settings, but no noticeable improvement is observed when adding policy information to the prompt.

Similar to the performance change observed between \citet{TEC_Sawhney} and \citet{sawhney-etal-2020-gpols}, all models except GPT-4o show decreased accuracy with the temporal split, but the overall trend of metadata enhancing performance persists.

\section{Analysis}
\begin{figure}
  \centering
  \includegraphics[width=0.5\textwidth]{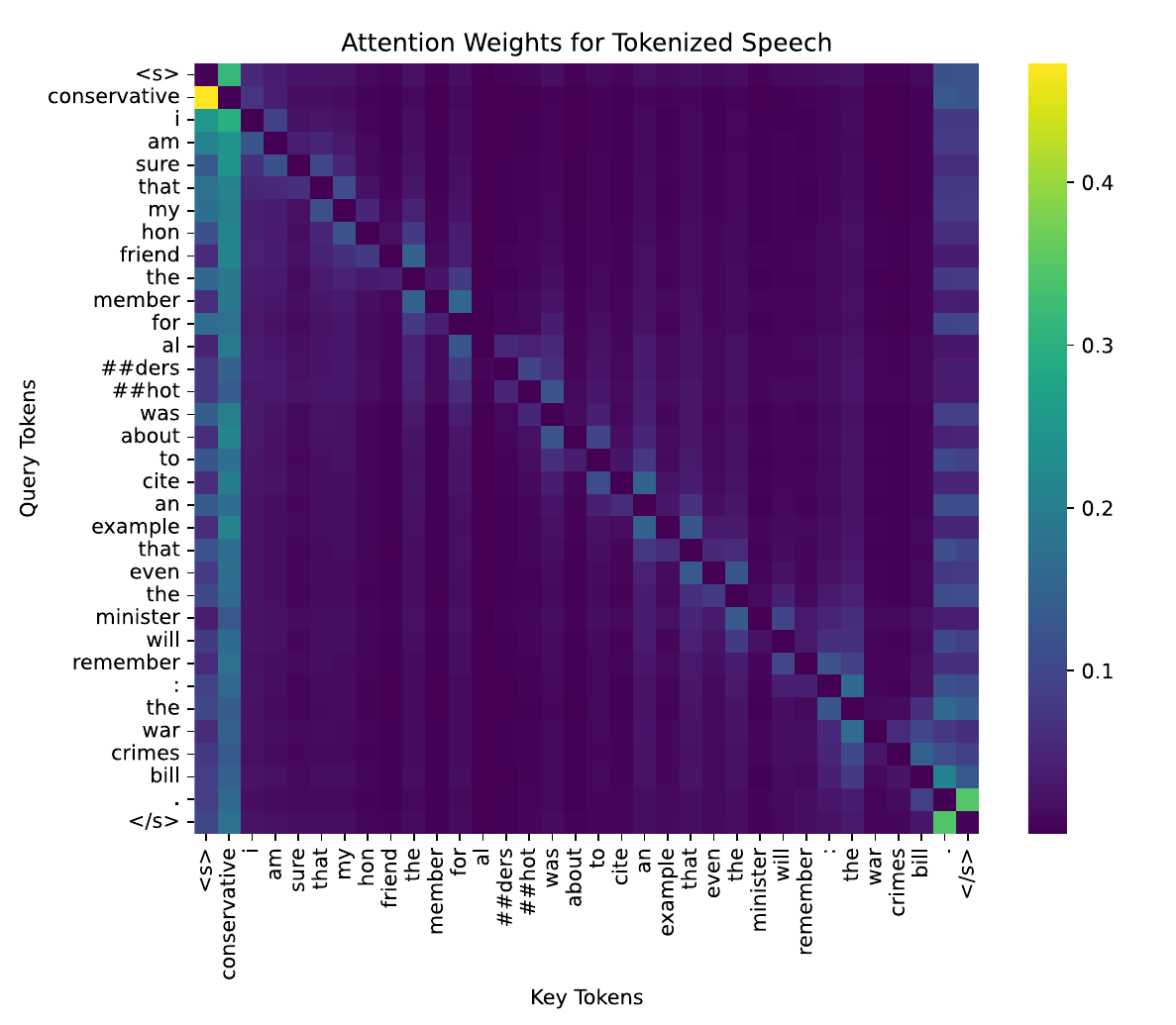}
  \caption{Attention weights averaged among all heads for a particular speech in response to a Labour party member. Speaker is a Conservative party member. Speech: ``\textit{I am sure that my hon friend the member for aldershot was about to cite an example that even the minister will remember : the war crimes bill.''}}
  \label{atten_party}
\end{figure}

The empirical evidence underscores the efficacy of a metadata-driven approach. Previous models have difficulty fully exploiting party and policy information, as evidenced by our party-based Naive Bayes model achieving state-of-the-art performance. Using only party information, we are able to achieve an accuracy of 0.80 on the full dataset.

Unlike prior work that employed more sophisticated methods, our results underscore the efficacy of a simple language model augmented with party and policy metadata. We find that when incorporating the party and policy metadata via probabilities as an extra feature to the language embedding, the classification layer outperforms our party-based Bayesian model, achieving an accuracy of 0.87. 

Augmenting the language model with metadata, MPNet consistently surpasses the Bayesian model in stance detection accuracy. Through a simple metadata prepending mechanism, MPNet performs better than concatenating engineered Bayesian probabilistic estimates into the language model. 
This is evident across various experimental settings, as shown in Figures \ref{model_acc_by_speech_len_PARTY} and \ref{model_acc_by_speech_len_PARTY_POLICY}. 
The overall trend favors the prepending mechanism, indicating its effectiveness in utilizing metadata even with minimal textual evidence. This trend is reinforced by Figure \ref{model_acc_by_uncertainty}, showing that as the prior becomes less uncertain, both models see performance gains.


In Figure \ref{atten_party}, we note that  the attention weights show that the party metadata is attended to throughout the entire speech. This highlights how the political party is important contextual information throughout the entire speech. A similar pattern is shown in Figure \ref{atten_policy_party} when policy metadata is prepended. Again, there is a consistent pattern of attention to the metadata throughout the entire speech. Notably, the high attention weight of the metadata query tokens on the BOS token suggests that MPNet has learned to distinguish this metadata as not part of the speech itself, but as a special token revealing information about the entire sentence.

GPT-4o is able to perform moderately well at this task.
Although many papers have shown that LLMs are good at zero and few-shot learning \citep{brown2020, lin-etal-2022-shot, frayling2024zeroshotfewshotgenerationstrategies,dabramo2024dynamicfewshotlearningknowledge}, from Table \ref{overall-stats}, we see that GPT-4o achieves an accuracy similar to those produced by baseline methods. GPT-4o also encounters difficulty when parsing complex political speeches, and further prompt refinement may increase performance.

\section{Conclusion}
Our findings reveal that simpler models can outperform more complex ones in political stance detection when utilizing the metadata properly. The success of metadata-enhanced language models not only establishes a new state-of-the-art on the \textit{ParlVote+} benchmark, but also invites further investigation into the properties of simpler language models and their attention mechanisms. This insight opens up new avenues for future research, particularly in examining how language models can interpret metadata to improve performance across a broader range of NLP tasks.

\section*{Limitations}
\subsection{Motion and Speech Truncation}
In this paper, we have limited the analysis of motions and speeches to a maximum length of 512 tokens, MPNet's maximum window size. This truncation method enables the model to grasp the basic stance presented in the texts. However, it's possible that significant details are omitted, which could potentially influence the accuracy of stance detection. Future iterations of this work might benefit from exploring alternative strategies to handle longer texts without losing pertinent information.

\subsection{Metadata Incorporation into Transformers}
The approach outlined in this paper incorporates Bayesian probabilities directly before the model's final classification layer, enhancing performance compared to using a metadata-free language model. Despite the success of this method, alternative integration techniques of Bayesian probabilities merit exploration. Investigating the effects of introducing these probabilities along with party and policy information at different stages of the model, particularly before the attention mechanisms, could reveal new insights. Additionally, the development of a more sophisticated final classification layer that operates on transformer-generated embeddings could offer further improvements.

\subsection{Additional Metadata}
Our research demonstrates the utility of prepending party and policy metadata to enhance model performance. Future studies could examine the impact of including metadata such as speaker name, date and time, and debate title, both in the initial model input and in generating aggregated statistics. This could involve calculating specific metrics, like a speaker's mean voting probability, that account for individual deviations from party lines, potentially introducing bias as a learnable parameter.

\bibliography{custom}

\appendix

\section{Appendix}
\label{sec:appendix}

\subsection{Incorporating Policy Information into the Bayesian Model}
\label{appdx:ttest}
To account for instances in which there are an insufficient number of examples to calculate a reliable value via equation \ref{naive_bayes_with_policy}, we conduct a 1-sample t-test to check if equations \ref{eqn:party} and \ref{naive_bayes_with_policy}
exhibit a statistically significant deviation. Thus, in our experiments that involve using the Bayesian model with policy data, we use equation \ref{naive_bayes_with_policy} if the null hypothesis of the t-test is rejected, which suggests that the policy incorporation significantly deviates from the party-only statistic. Otherwise, we use the value from equation \ref{eqn:party}. Intuitively, we attempt to obtain a probability at the policy level, defaulting to the global party-based statistic in the case where we detect no statistically significant deviation between the two.

\subsection{Optimization}
For all experiments in this paper, we employ the AdamW optimizer with the default hyperparameters \citep{loshchilov2019decoupled}.

\subsection{Technicality on Maximum Token Size}
Due to the maximum token size for language models, directly concatenating the motion with the transcript results in significant truncation of the transcript, as they are much longer on average than the motions. This makes classification difficult, as important parts of the speech transcript are truncated. To account for this, we run the language model over the motion and transcript in sequence, and concatenate the resulting embeddings. We then pass this through a fully-connected layer with a non-linear activation function to create the resulting 2-dimensional encoding. 

\subsection{Llama 3 8B Results}
Below, we include our initial experimentation with using Llama 3 8B, though we believe future work with the instruction-tuned version should be done to validate these results.
\begin{table}[h]
\footnotesize
\centering
\begin{tabular}{lcccc}
\toprule
\multicolumn{5}{l}{\textbf{Pure Language Models}} \\
Model & Text & \makecell{Parties} & \makecell{Policies} & Acc.\\
\midrule
Llama3 8B (0-shot) & \ding{51} & \ding{55} & \ding{55} & 0.49 \\
Llama3 8B (5-shot) & \ding{51} & \ding{55} & \ding{55} & 0.51 \\
Llama3 8B (5-shot) & \ding{51} & \ding{51} & \ding{51} & 0.51 \\
\bottomrule
\end{tabular}
\caption{Selected Llama 3 8B results on random data split}
\end{table}

\subsection{GPT-4o prompts}
\label{chatgpt_prompts}
This section shows the prompts we used for evaluating GPT-4o on the \textit{ParlVote+} dataset. We include in this section the system and user prompts, as well as the format for the few-shot examples and metadata that were presented to the model.
\subsubsection{System Prompt}
\begin{spverbatim}
You are a classification model that is really good at following instructions and produces brief answers that users can use as data right away. Please follow the user's instruction as precisely as you can.    
\end{spverbatim}

\subsubsection{Few-shot Example Format}
\begin{spverbatim}
[METADATA]

Motion: [MOTION]

Speech: [SPEECH]

Correct Answer: [0 or 1]
\end{spverbatim}

\subsubsection{Metadata Format}
\begin{spverbatim}
Party of Motion: [MOTION PARTY]

Party of Speech: [SPEAKER PARTY]

Policy: [POLICY]
\end{spverbatim}

\subsubsection{User Prompt 0-shot}
\begin{spverbatim}
You will be presented with a motion and a speech from different representatives in the UK Parliament. Your task is to classify whether the speech supports or does not suppor the motion. Please respond with a 0 if the speech does not support the motion and a 1 if the speech does support the motion.\\

[METADATA]

Motion: [MOTION]

Speech: [SPEECH]
\end{spverbatim}

\subsubsection{User Prompt 6-shot}
\begin{spverbatim}
You will be presented with a motion and a speech from different representatives in the UK Parliament. Your task is to classify whether the speech supports or does not support the motion. Please respond with a 0 if the speech does not support the motion and a 1 if the speech does support the motion.

Here are some examples, where you are presented the motion, then the speech, and finally the correct answer which is either a 0 or a 1:

[EXAMPLE 1]

...

[EXAMPLE 6]

Now, please classify the following motion and speech with a 0 if the speech does not support the motion and a 1 if the speech does support the motion.

[METADATA]

Motion: [MOTION]

Speech: [SPEECH]
\end{spverbatim}

\subsection{Additional Figures}
\begin{figure*}
  \centering
  \includegraphics[width=\textwidth]{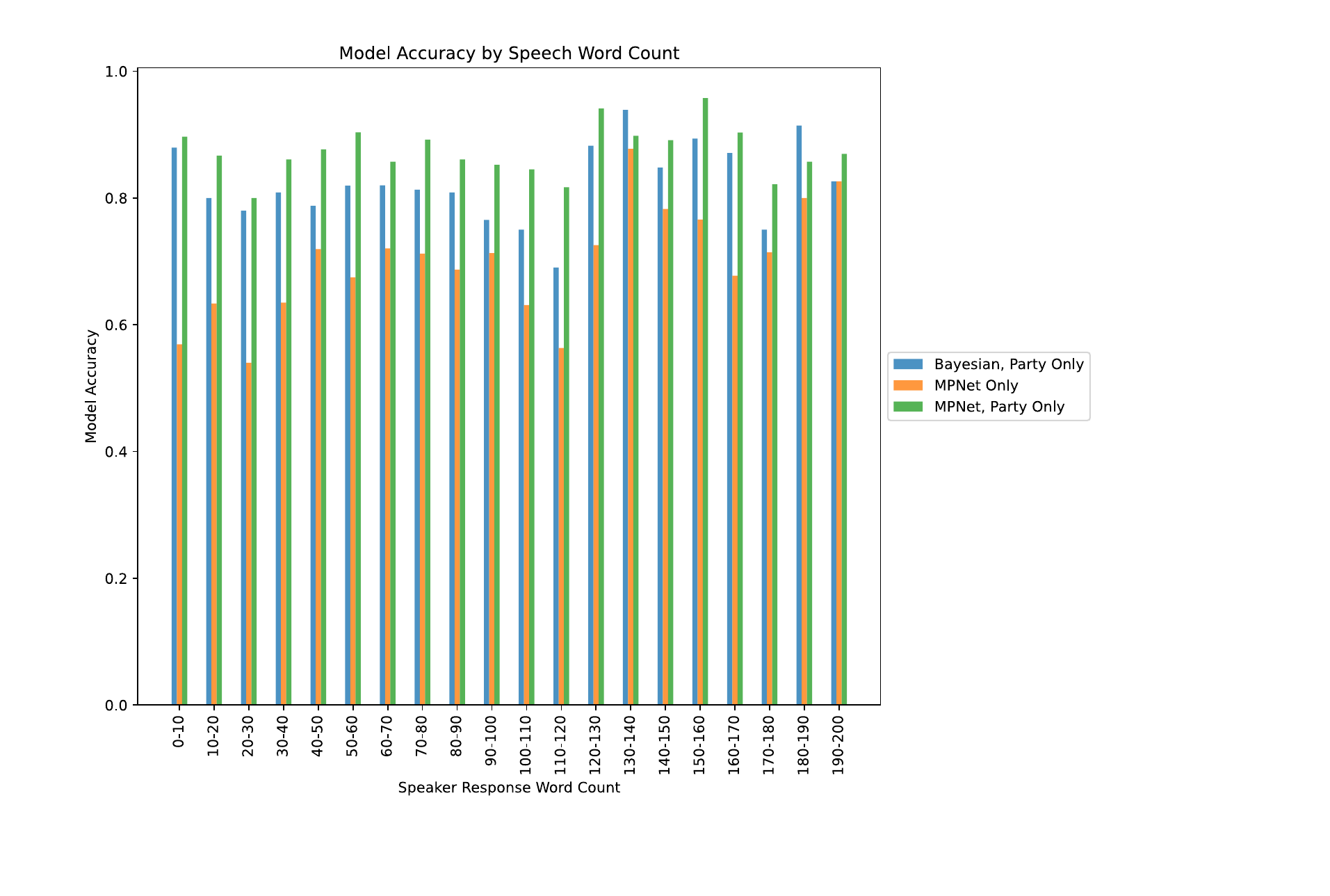}
  \caption{Model Accuracy by Speaker Response Length / Word Count (with only party information)}
  \label{model_acc_by_speech_len_PARTY}
\end{figure*}

\begin{figure*}
  \centering
  \includegraphics[width=0.8\textwidth]{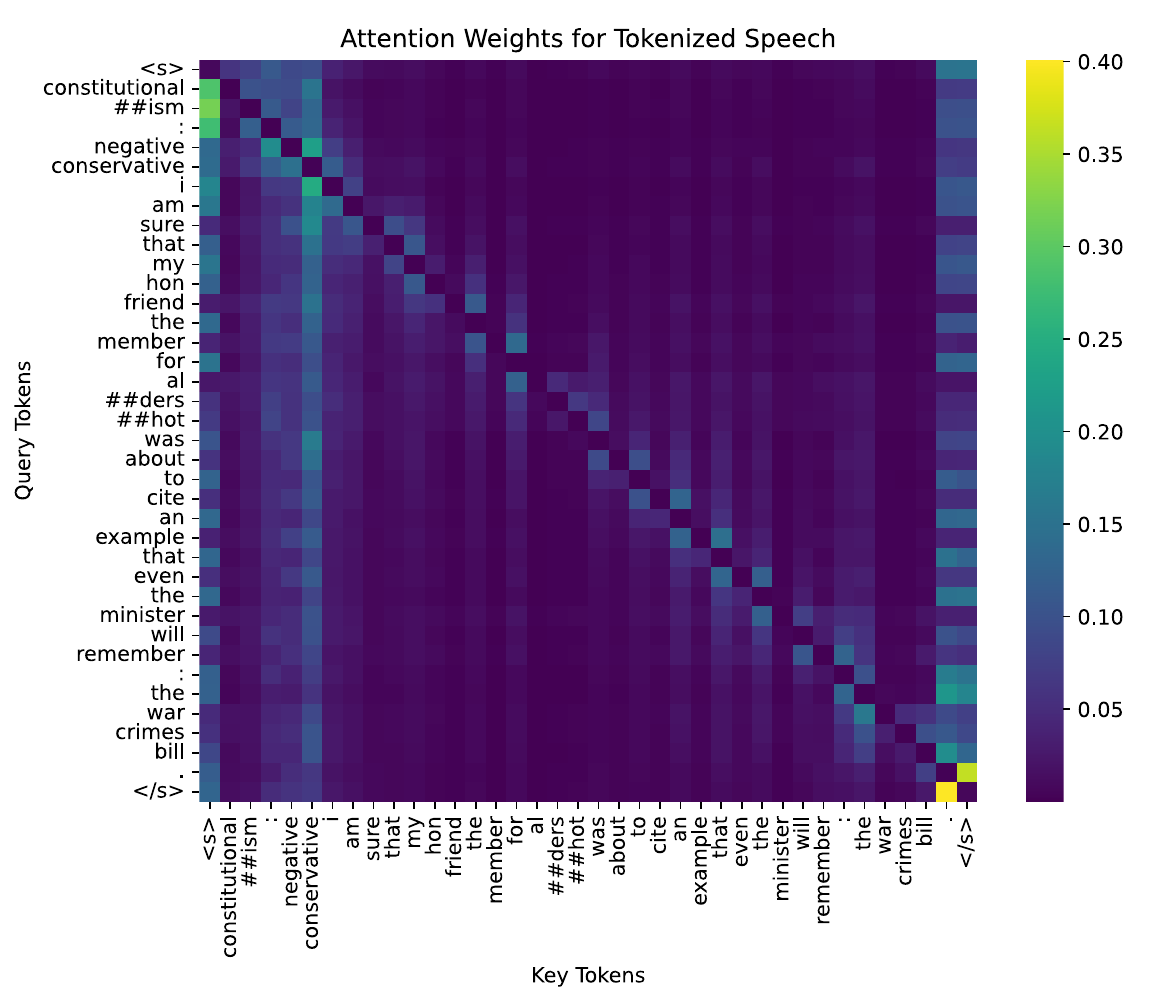}
  \caption{Same speech as Figure \ref{atten_party}. The policy in question is ``Constitutionalism: Negative.''}
  \label{atten_policy_party}
\end{figure*}

\begin{figure*}
  \centering
  \includegraphics[width=\textwidth]{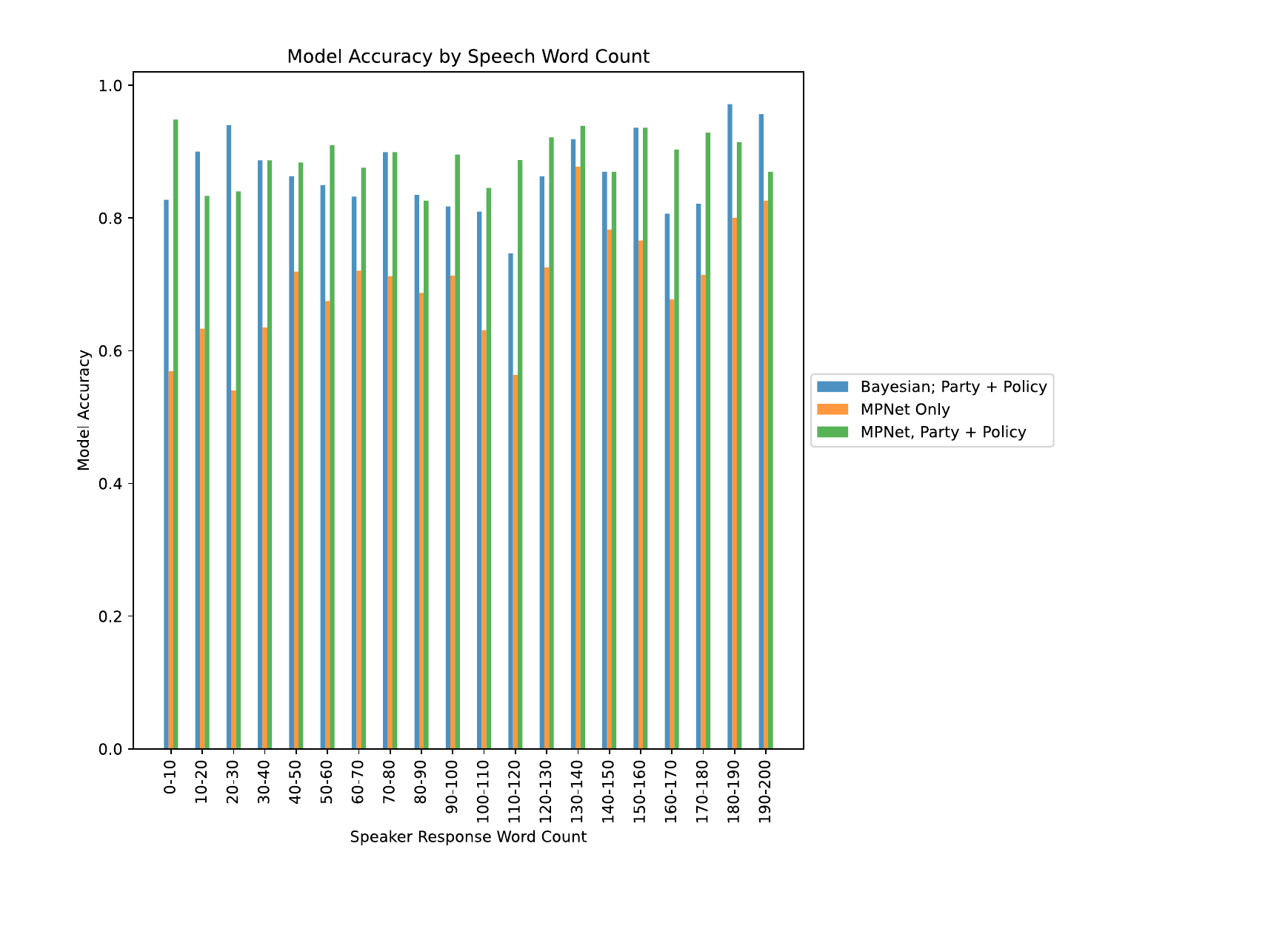}
  \caption{Model Accuracy by Speaker Response Length / Word Count (with party and policy information)}
  \label{model_acc_by_speech_len_PARTY_POLICY}
\end{figure*}

\begin{figure*}
  \centering
\includegraphics[width=0.80\textwidth]{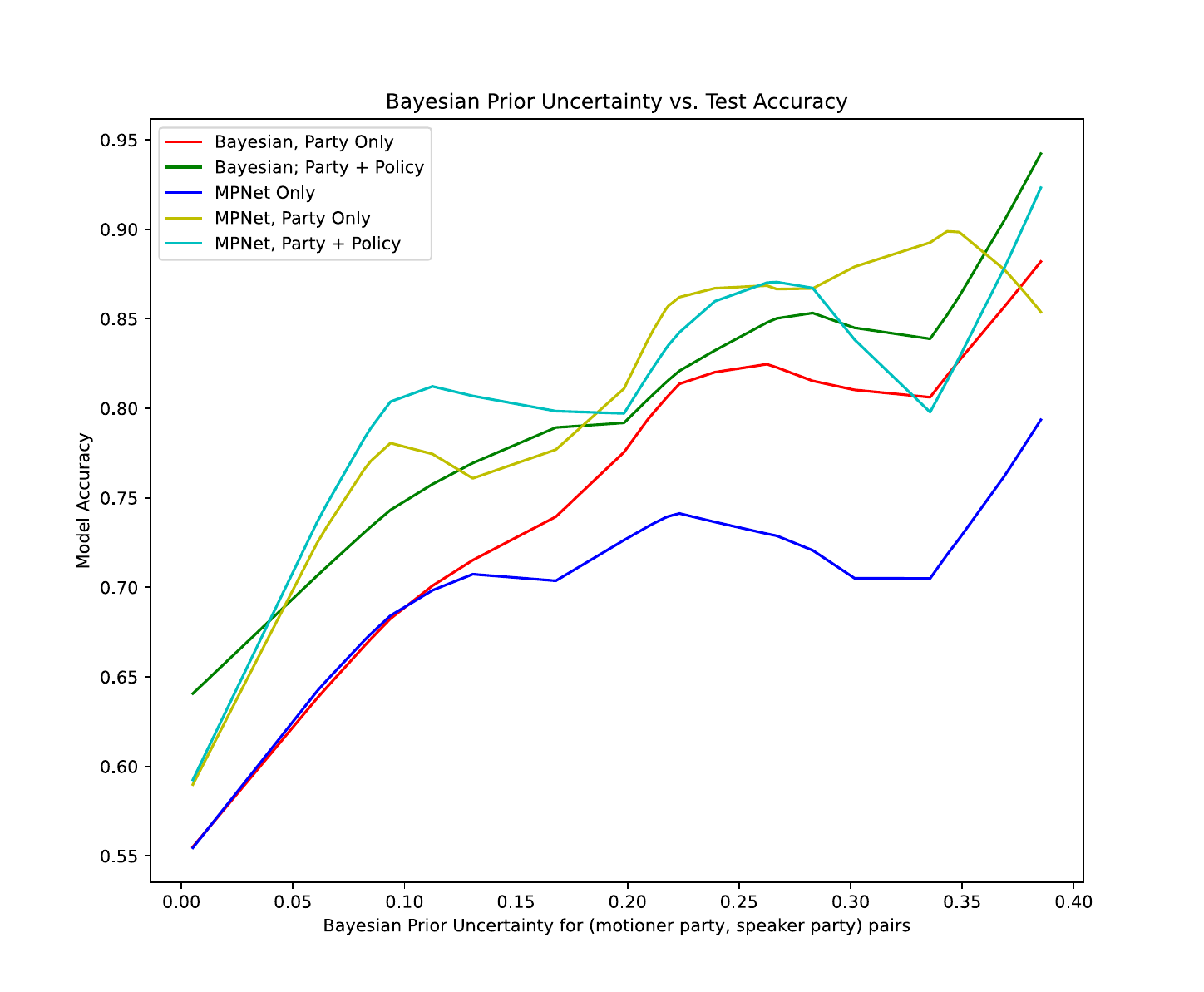}
  \caption{Model Accuracy by Bayesian Prior Uncertainty $\left(|p - 0.5|\right)$ for different motioner party + speaker party pairs. To prevent the effects of outliers, we only include datapoints where the number of examples exceeds 50. We find the curve of best fit using Locally Weighted Scatterplot Smoothing (LOWESS)}
  \label{model_acc_by_uncertainty}
\end{figure*}


\end{document}